\documentclass[conference]{IEEEtran}
\IEEEoverridecommandlockouts

\usepackage{cite}
\usepackage{amsmath,amssymb,amsfonts}
\usepackage{algorithmic}
\usepackage{graphicx}
\usepackage{textcomp}
\usepackage{algorithm}  
\usepackage{xcolor}
\usepackage{verbatim}
\usepackage{float}
\usepackage[bookmarks=false]{hyperref}
\usepackage{booktabs}
\usepackage{color}

\def\BibTeX{{\rm B\kern-.05em{\sc i\kern-.025em b}\kern-.08em
    T\kern-.1667em\lower.7ex\hbox{E}\kern-.125emX}}
\begin{document}

\title{Fast and Robust Bin-picking System for Densely Piled Industrial Objects*\\
\thanks{*This work is supported by Key Research and Development Program  of Zhejiang Province of China under Grant 2020C01114.}
}

\author{\IEEEauthorblockN{Jiaxin Guo,
Lian Fu, Mingkai Jia,
Kaijun Wang, Shan Liu}
\IEEEauthorblockA{\textit{State Key Laboratory of Industrial Control Technology,} \\
\textit{College of Control Science and Engineering} \\
\textit{Zhejiang University}\\
Hangzhou, Zhejiang, 310027, China \\
Email{\{jiaxinguo, lian.fu, mingkaijia, kjwang, sliu\}@zju.edu.cn}}}

\maketitle
\begin{abstract}
Objects grasping, also known as the bin-picking, is one of the most common tasks faced by industrial robots. While much work has been done in related topics, grasping randomly piled objects still remains a challenge because much of the existing work either lacks in robustness or costs too much resource. In this paper, a fast and robust bin-picking system is developed to grasp densely piled objects adaptively and safely. The proposed system starts with point cloud segmentation using improved density-based spatial clustering of applications with noise (DBSCAN) algorithm, improved by combining the region growing algorithm as well as using Octree to speed up the calculation. The system then uses principle component analysis (PCA) for coarse registration and for fine registration, the iterative closest point (ICP) algorithm is used. We propose grasp risk score (GRS) to evaluate each object by the collision probability, the stability of object and the whole pile's stability. Through real test with Anno robot, our method is verified to be advanced in speed and robustness.
\end{abstract}


\section{Introduction}
Industry has seen a rapid growth of robots application. Humdrum works, such as bin-picking, met the first wave of robots in job. Traditional bin-picking robots have limited ability sensing the environment, making it hard to meet the requirements of the complicated real-situation in plants. Recent studies focus on improving visual approaches of robots. Point cloud, with its richness in 3D information, is born to be among the best choices for object sensing. Generally, bin-picking problems are solved by a common pipeline, which basically contains plane segmentation, correspondence grouping, coarse registration and fine registration.

Previous researches about pipeline usually focus on some specific procedures. Random sample consensus (RANSAC) works well in solving plane segmentation problems, and some more modified algorithms are proposed recently. Bastian Oehler et al. present a method that can segment 3D point clouds into planar components efficiently with multi-resolution, which combines the Hough transform with robust RANSAC\cite{oehler2011efficient}. Lin Li et al. propose an improved RANSAC method to avoid spurious planes for 3D point-cloud plane segmentation on the basis of Normal Distribution Transformation (NDT) cells\cite{li2017improved}. Credit to Xian-Feng Han et al., filter algorithms are categorized into four classifications\cite{han2017review}, statistical-based\cite{narvaez2006point}\cite{schall2005robust}, neighborhood-based\cite{hu2006mean}\cite{wang2013consolidation}, projection-based\cite{ye2011accurate} and PDEs-based filtering\cite{taubin1995estimating}\cite{xiao2006dynamic}. Sample consensus initial alignment (SAC-IA) is one of the basic coarse registration approaches, with different modified versions specifically designed for different situations\cite{chen2017point}\cite{rusu2009fast}. Besides, S Zhang et al. use frequency domain point cloud registration based on the Fourier transform to detect the side data of the whole part\cite{zhang2019frequency}. For fine registration, a few new approaches are proposed recently. Weimin Li et al. propose a modified iterative closest point (ICP) algorithm with high speed\cite{li2015modified}. Brayan S Zapata-Impata et al. propose a method to find the best pair of the three-dimensional grasping points for the unknown objects with a partial view. In order to perform grasps autonomously, the robot needs to know where to place the robotic hand to ensure stable grasps.\cite{zapata2017using}. For object pose estimation, modified algorithms like adaptive threshold for bin-picking are proposed\cite{yan2020fast}.This pipeline as a whole has also been covered by a handful of researches. Dirk Buchholz et al. show an applicable solution for the bin-picking problem, focusing on robustness against noise and object occlusions\cite{buchholz2010ransam}\cite{buchholz2013efficient}. Ales Pochyly et al. present a functional case study, which concentrates on the bin-picking system on the basis of the modified revolving vision system, in which they admit it remains hard to implement in industry environment and give a few analysis\cite{pochyly2017robotic}. While above studies are verified under simulation, there are some papers give pretty results using real robot. Carlos Martinez et al. prove their pick-policy and basic visual system to be robust using ABB IRB2400 robot, and give a practical test method\cite{martinez2015automated2s}. In addition to eye-to-hand robots, eye-in-hand is also capable to deal with bin-picking problem according to Wen-Chung Chang’s vision-based robotic bin-picking system in which CCP approach is proposed\cite{chang2016eye}. Some patents are also applied, for example, Kye-Kyung Kim’s patent about bin-picking system using top-down camera with bin-picking box\cite{kim2020bin}.

However, though much work has been done in bin-picking, regarding the problem of randomly-piled-objects bin-picking problem which is a more general scenario in real industry application, only a few papers have been published. One example is inward-region-growing-based accurate partitioning of closely stacked objects for bin-picking proposed by Zaixing He et al. Focusing on segmentation of piled objects, they give no result on real-robot\cite{he2020inward}. 

We make two main contributions in this paper: 
\begin{itemize}
\item An Improved density-based spatial clustering of applications with noise (DBSCAN) segmentation algorithm is designed.
\item An efficient and adaptive grasp-policy evaluation function is proposed.

\end{itemize}

The rest of the article is organized as follows. Section 2 presents an overview of our point-cloud-based bin-picking algorithm. In Section 3, our methods are described in detail. And then in Section 4, results and discussion of experiments are illustrated. Conclusions are drawn in Section 5.

\section{Overview}

\begin{figure}[ht]
\centering
\includegraphics[width=0.48\textwidth]{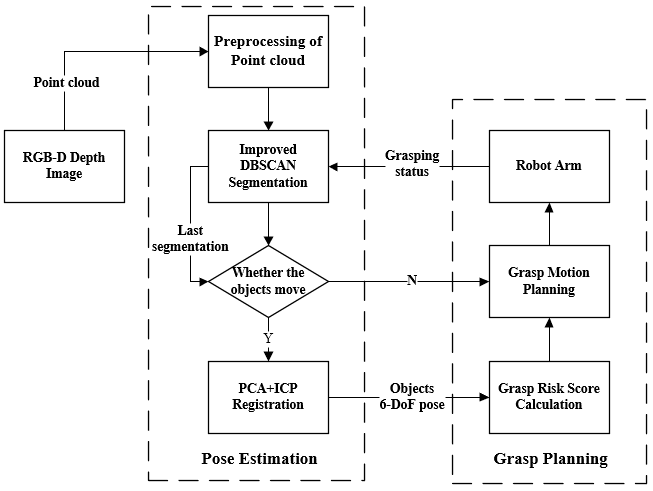}
\caption{The schematic diagram of our approach.}
\label{pipeline}
\end{figure}

The goal of our research is to develop a fast bin-picking system for grasping dense and stacked objects adaptively, safely and fast. Here we outline the given information, the potential challenges and the desired capabilities of the proposed method.

Our system is expected to get depth images from RGB-D camera and convert it into 2.5D point cloud for pose estimation. 2.5D point cloud has insufficient dimensionality for the single viewpoint of camera. As we focus on grasping general industrial objects like three-way-pipes etc. which have similar textures and features, the objects' computer-aided design models with their 3D point cloud can be assumed as priori knowledge. Simple as their shapes are, it becomes a big challenge to estimate their 6-DoF pose accurately when they are randomly and densely piled. In that case, some traditional segmentation methods are not suitable and robust for this scene, some feature-based registration methods are accurate but slow and some deep learning methods are robust but have too much calculating and time cost.

Our approach is shown in Fig.\ref{pipeline}. We propose a DBSCAN-based point cloud segmentation algorithm improved through combining the Region Growing algorithm and using Octree to speed up the calculation. Our segmentation algorithm is verified to perform well in dense point cloud segmentation and noise point cloud filtering. To reduce cost in features calculation, we use principle component analysis (PCA) to estimate the principle component of the segmented point cloud for coarse registration, then for fine registration, ICP algorithm is used. For the safety goal, we focus on a novel policy of grasping for the objects to avoid the collision while improve the grasping efficiency. We consider three factors to evaluate the collision probability, the stability of object and the whole system. According to the factors, we define the grasp risk score (GRS) to evaluate each object. From the rank of GRS, we get the optimal sequence of grasping. In the end of a grasping round, the program will check if the objects moved by the change of segmented centroids. The program will do registration and calculate GRS again only when a significant displacement has occurred, otherwise will continue grasping without data process to improve efficiency.

\section{Methods}
\subsection{Preprocessing}\label{AA}

In our approach, after using RGB-D camera to get point cloud of grasping area, it is necessary to preprocess the point clouds so that we can remove the noise and get the point cloud of target objects. The implementation procedure of preprocessing is as following:
\begin{itemize}
\item Convert the depth image to point cloud.
\item Extract the grasping area and remove the noise by using the passthrough filter.
\item Use voxel grid filter to downsample, that is to reduce the number of point cloud but retain the detailed information of original point cloud.    
\item Use RANSAC to remove the plane of the point cloud rapidly\cite{schnabel2007efficient}.
\end{itemize}

\begin{figure}[ht]
\centering
\includegraphics[width=0.45\textwidth]{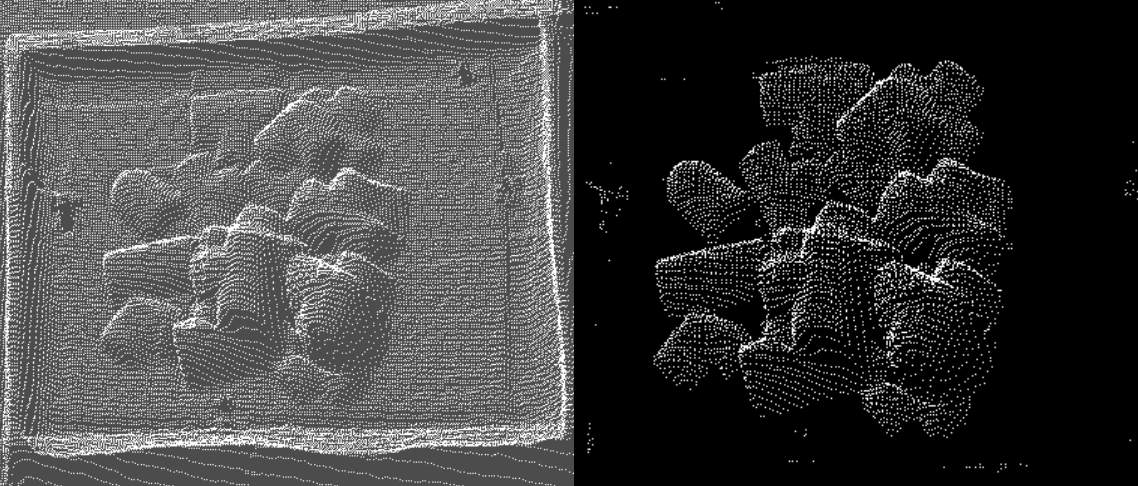}
\caption{Preprocessing result. }
\label{pre}
\end{figure}

The preprocessing result is shown in Fig.\ref{pre}.
\subsection{Improved DBSCAN algorithm for segmentation}
There has been many point cloud segmentation algorithms, like Region growing, Euclidean Clustering, RANSAC and so on. However, they are not robust for piled and dense objects, which have arbitrary poses and overlapping point cloud. To overcome this problem, the paper proposes an improved DBSCAN algorithm for point cloud segmentation of stacked and dense objects, combining Region Growing with DBSCAN and using Octree to improve the efficiency of algorithm.
\par
DBSCAN algorithm is a simple and effective clustering algorithm on the basis of density. It can seek out arbitrary shape clustering as well as remove noise without any prior knowledge\cite{ester1996density}. There are 2 parameters required for DBSCAN, $Eps$ and $MinPts$. Firstly, DBSCAN evaluates the points density by defining a cluster as the maximum set of points whose neighbor points are within $Eps$. Secondly, $MinPts$ is the minimum number of neighbors of the core points forming the clusters. The border points are contained in the Eps-neighborhood of the other core points. The noise points are the points that not belonging to any cluster.
\par
Here are some important definitions of DBSCAN.
\par
\textbf{Definition 1:} (directly density-reachable) It is directly density-reachable between a pair of points p and q wrt. Eps, MinPts if:
\begin{gather}\label{densityreachable}
	p \in N_{Eps} (q) \\
|N_{Eps}(q)| \geq MinPts
\end{gather}
\par
Directly density-reachable is symmetric for pairs of core points, but not for the pair of a core point and a border point.

\par
\textbf{Definition 2:} (density-reachable) It is density-reachable between a pair of points p and q wrt. Eps and MinPts if there is a chain of points $p_1, p_2, ..., p_n, p_n=p, p_1=q$, in which point $p_{i+1}$ is directly density-reachable from $p_i$.
\par
The density-reachability is typically extended from the direct density-reachability. The relationship between them is transitive but not symmetric.
\par
\textbf{Definition 3:} (density-connected) It is density-connected between a pair of points and q wrt. Eps and MinPts if there is a point o, from which both p and q are density-reachable wrt. Eps and MinPts.
\par
As a symmetric relation, density-connectivity is also reflexive for density-reachable points .
\par
\textbf{Definition 4:} (cluster) Let D represent for a dataset of points, of which the non-empty cluster C wrt. Eps and MinPts is a subset that meets the conditions below:
\par
1)$\forall p$, $q$: if $p \in C$ and $q$ is density-reachable from p wrt. Eps and MinPts, then $q \in C$.(Maximality)
\par
2)$\forall p$, $q \in C$: p is density-connected to q wrt. Eps and MinPts.(Connectivity)
\par
\par
For cluster finding, DBSCAN will start at a point p, which is arbitrarily selected and all points  wrt. Eps and MinPts density-reachable from p are retrieved. For the core point p, DBSCAN generates a cluster wrt. Eps and MinPts, and for the border point p, it will visit the next point. 
\par
There are some deficiencies of DBSCAN. First, it costs a lot when finding the neighbor points to form clusters. It's necessary to speed up the calculation to improve the efficiency. Second, DBSCAN algorithm only merges the Eps-neighborhood points of core points, which means lots of border points are lost in this proceeding. It causes a loss of the objects' edge and some detailed point cloud, which may lead to a bad registration of following proceedings.  
\par
\textbf{Improved DBSCAN: }
\par
1) Use Octree to speed up:
\par
To improve the efficiency of DBSCAN for finding the nearest neighbors, it is necessary to use relevant algorithms to speed up the calculation. As a dynamic partition algorithm, K-d tree is efficient and has been widely implemented for handling point cloud. However, all of the child nodes should be retrieved to traverse the child nodes where the 3D boundary satisfies a positional query. Inspired by Soohee Han's work\cite{han2018towards}, Octree is adopted in this paper.
\par
Octree is a tree used to describe three-dimensional space. Each node of octree represents a cube volume element, and each node has eight child nodes. When using Octree for point cloud, the 3D boundary is divided into eight octants, which will be further subdivided recursively only when they bear points within themselves until the sequence reaches a given threshold value $depth$. In Octree, only one child node in each depth is traversed for the 3D boundary of each node is known by positional query. And a leaf node can be advantageously retrieved to improve the index efficiency.
\par

By using Octree to find the nearest points, the accuracy and efficiency of DBSCAN are improved to deal with 3D point cloud data.
\par
2) Combine with Region Growing algorithm
\par
To avoid the loss of some details and bounds, we present a approach combining the DBSCAN with Region Growing algorithm\cite{rabbani2006segmentation}. Firstly, Region Growing sorts the points by their curvature value and picks points with minimum curvature value as seeds for the region's growth. Secondly, the Region Growing algorithm measures local geometry by fitting a plane to some point's neighborhood to find surface normals. Using the residual of plane fitting, the local curvature is also approximated. Two parameters $\theta_{th}$ and $r_{th}$ are involved in the method, which mean curvature threshold and smoothness threshold. According to the two parameters, Region Growing focuses on using the points' normal and residuals to group points with smooth surfaces.
\par
DBSCAN usually results in over segmented point clouds and lost borders, but Region Growing tends to extend for more points from the growing seeds. To enhance the DBSCAN algorithm for solving the problems, we combine the Region Growing method with DBSCAN algorithm, which can take DBSCAN clusters as region growing seeds to start growing.  
Instead of only merging the clusters of core points into one, we also merge the border point cluster to the core point cluster according to some rules: 
\begin{itemize}

\item The border point and core point are directly density-reachable.
\item The Eps neighborhoods of two points have the most in common.
\item The normal of a point is estimated using it's neighboring points to fit a plane. The curve features and normal of two pofts' Eps neighborhood can be matched, which means the two clusters have similar surface properties.
\end{itemize}
\par

\begin{figure}[ht]
\centering
\includegraphics[width=0.48\textwidth]{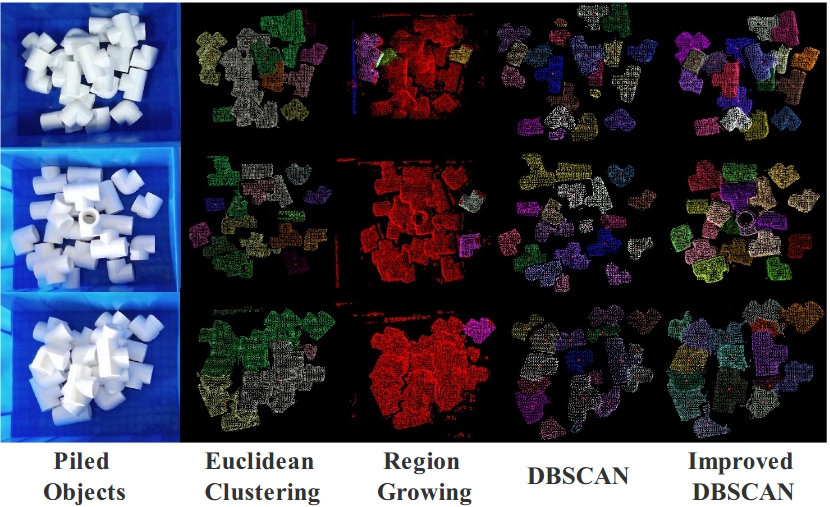}
\caption{Segmentation results of different algorithms. The centroids of
each cluster are calculated and shown as red points. }
\label{seg}
\end{figure}

\begin{algorithm}[!h]
	\caption{Improved DBSCAN}
	\label{dbscan}
	\begin{algorithmic}[1]
	\STATE input: Octree depth $d$, $Eps$, $MinPts$, Point cloud = $\left\{ P \right\}$, Point normals = $\left\{N\right\}$, Point curvatures = $\left\{c\right\}$, smooth threshold $\theta_{th}$, curvaturethreshold $c_{th}$
		\STATE Use Octree to voxelize the point cloud with depth less than $d$.
		\STATE Use Octree to find $Eps$-neighborhood of each points based on Euclidean distance. 
		\STATE According to $MinPts$, find the core points and border points, exclude the noises.
		\STATE Merge the clusters according to Definition 4.
		\STATE For border points, find their directly density-reachable core points.
		\STATE Find the cluster of the core points which have most in intersaction.
		\STATE Calculate the normals and curvature of core point cluster and border point cluster, if less than $c_{th}$ and $\theta_{th}$, merge them into new clusters.
		
	\end{algorithmic}
\end{algorithm}

The algorithm proceeding is in algorithm \ref{dbscan}. The segmentation result is shown in Fig.\ref{seg}. According to Fig.\ref{seg}, for Euclidean Clustering and Region Growing, the segmentation could not separate the dense and overlapping point cloud of each object accurately, and some noise could not be removed; for DBSCAN, the segmentation separates well but loses details and borders of objects; for our method, the above objects which have relatively complete point cloud are segmented, and some bottom objects and noise point cloud are filtered.

\subsection{Registration}

After segmentation of piled objects, a fast and robust registration method is necessary for estimate the pose of objects. This paper fulfills the coarse registration with PCA and uses ICP to further improve the accuracy.

\textbf{Coarse Registration:} To reduce the dimensions of the data set, PCA algorithm retains the greatest feature of the data which contribute to variance at most. For point sets $P(x_1,x_2, ...x_n)$, $x_i$ is n-dimensions data, and the algorithm calculate the mean and variance\cite{yuan20163d}, as shown in \ref{PCA} and \ref{PCA2}.

\begin{gather}
\overline{x} = \frac{1}{n}\sum_{i=1}^n x_i\label{PCA} \\
covP = \frac{1}{n}\sum_{i=1}^n (x_i-\overline{x})(x_i-\overline{x})^T\label{PCA2}
\end{gather}
\par
The principle component of the point cloud P can be figured out from the co-variance matrix's feature vectors $covP$. For the point cloud data which has three dimension, here we denote $overline{x}$ as the coordinate system's origin, and use the three feature vectors as the XYZ axes which represent the principle component of the data. The coarse registration can be achieved by adjusting the coordinate systems between the model point cloud and the target point cloud.

\textbf{Fine Registration:}
Point-to-point ICP are applied for fine registration to estimate the pose of object, thus improving the accuracy of registration. ICP needs an iterative initial transformation matrix T, which can be set by the result of coarse registration. Then point-to-point ICP updates the transformation T through minimizing the objective function $E(T)$:
\begin{gather}\label{ICP}
E(T) = \sum_{(p,q) \in D} ||p-Tq||^2
\end{gather}

\begin{figure}[ht]
\centering
\includegraphics[width=0.45\textwidth]{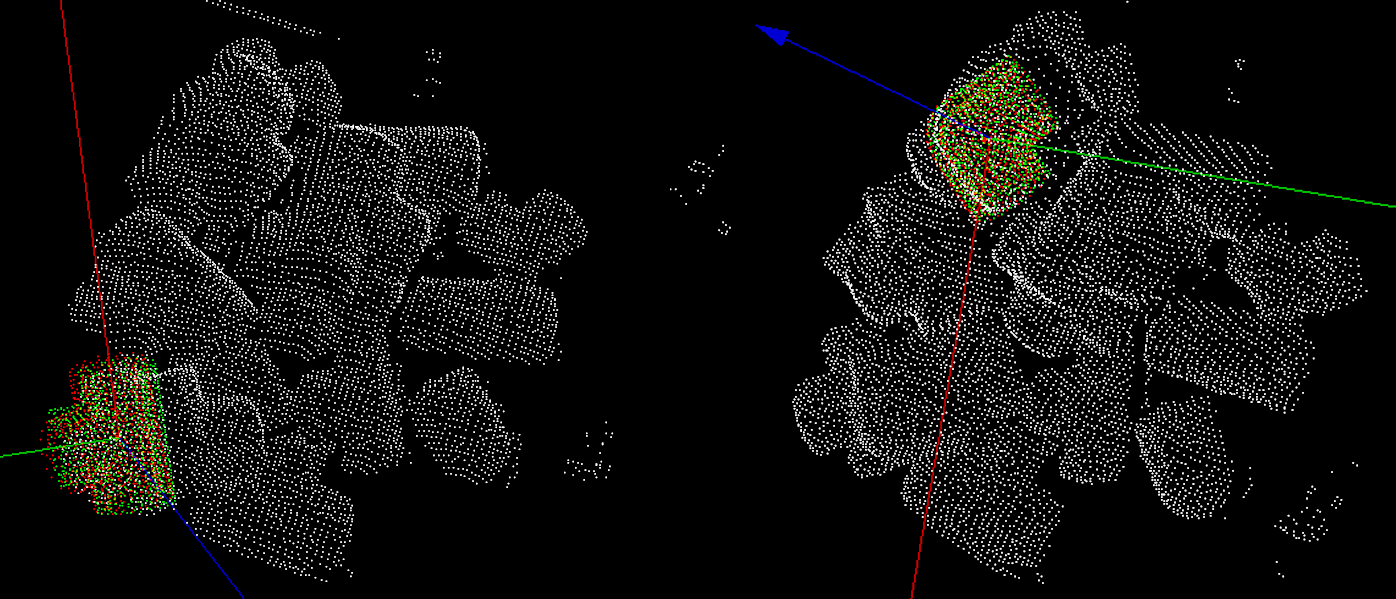}
\caption{Registration results. The principle components are shown as three axises.}
\label{regi}
\end{figure}

The registration results are shown in Fig. \ref{regi}.

\subsection{Policy of grasping}
Grasp policy involves grasp order, grasp position, grasp angle and gripper opening width. Inspired by Kentaro Kozai's work\cite{kozai2018determining}, we propose Grasp Risk Score(GRS) to optimize above parameters in case to avoid collision and improve the grasping efficiency. 

\subsubsection{Grasp order selection}

We consider three factors in grasp order selection. $P_{self}(i)$ represents the unstability of object $i$ itself. We use $P_{grasp}(i)$ to describe the possibility of collision while grasping object $i$. $P_{system}(i)$ tells the contribution to the system's stability of object $i$, that is, if the system would tend to collapse without object $i$, then the $P_{system}(i)$ would be pretty large in value.

We define the Grasp Risk Score(GRS) of object $i$ in equation \ref{GRS}.
\begin{equation}
	GRS(i)=(\alpha P_{self}(i)+\beta P_{grasp}(i))\cdot P_{system}(i) \label{GRS}
\end{equation}
with $\alpha,\beta$ to be tuned.

The GRS expresses the possibility of object $i$ to be safely picked up by formulating the Grasp Policy as a function of the grasping order. The optimal grasping parameter $\tilde{i}$ can be determined by maximizing GRS, as in equation \ref{i_optimal}.

\begin{equation}
	\tilde{i}=\mathop{\arg\min}_{i \in P}GRS(N) \label{i_optimal}
\end{equation}
with $N$ being the number of objects.

$P_{self}(i)$ expresses the unstability of the object $i$ itself. We consider the height of the object $i$'s barycenter $z_{oi}$ and the principal axis angle $\theta_i$ to be the main factors that determine the stability of the object. Generally, a large $z_{oi}$ means object $i$ is on the top of the pile, while a small one means it's lying on the ground, which is obviously more stable. And a slant principal axis means it tends to topple over. $P_{self}(i)$ is defined as equation \ref{Pself}

\begin{equation}\label{Pself}
	P_{self}(i)=k_1\cdot z_t+k_2\cdot \sin{\theta}
\end{equation}
with $k_1$ and $k_2$ to be tuned.

$P_{grasp}(i)$ represents the possibility of collision while grasping object $i$. Since the grasp parameters are uncertain, we just calculate the density of objects around object $i$ by using KD-Tree to find the distance of K nearest neighbors as an estimate. Denote the centroid position of object $i$ as $(x_i,y_i,z_i)$, the centroid position of other object as $(x_j,y_j,z_j)$. $P_{grasp}(i)$ is defined as formula \ref{PD}.

\begin{equation}\label{PD}
	P_{grasp}(i)=\frac{1}{K}\sum \limits_{i \in K}\{(x_i-x_j)^2+(y_i-y_j)^2+(z_i-z_j)^2\}
\end{equation}

$P_{system}(i)$ expresses object $i$'s contribution to the system's stability. An object can count a lot if it's supporting or propping against another object. As point cloud data is gained from an overhead camera, it contains only the information of the top layer of the object pile, thus above situations would be presented as overlap. Through advanced segmentation and registration, the dependencies information could be gained. Replace the origin point cloud $Q_i$ with the template point cloud $\tilde{Q_i}$ introduced during the registration. For target object $O_{target}$, we have formula \ref{system}.
\begin{equation}\label{system}
P_{system}(i)=\left\{
\begin{array}{rcl}
0 & & {threshold < \mathop{COUNT}\limits_{i\in N}(\tilde{Q}_{target}\cap\tilde{Q_i})}\\
1 & & {threshold \geq \mathop{COUNT}\limits_{i\in N}(\tilde{Q}_{target}\cap\tilde{Q_i})}
\end{array} \right.
\end{equation}
with $N$ being the number of detected objects and $COUNT(Q)$ meaning count the points' number of point cloud $Q$.

\subsubsection{Grasp position determination}

As Brayan S et al. proposed, the grasp point should on the object's principle axis $l_{axis}$\cite{zapata2017using}. Besides, the rules exists that the smaller the distance from the center of mass the better, and the larger the distance from the nearby object to the barycenter the better. Denote the distance from the lower endpoint of the principle axis to the grasp point as $x$, the distance between the barycenter and the grasp point as $de(x)$, the distance between the grasp point and the nearest object as $dn(x)$. The grasp point's position $\tilde{x}$ is determined by formula \ref{point}.

\begin{equation}\label{point}
	\tilde{x}=\mathop{\arg \min}\limits_{x} k_3\cdot de(x)+k_4\cdot dn(x)
\end{equation} $k_3$ and $k_4$ to be tuned.

\section{Experiments and Results}
We evaluate the proposed algorithm's performance and and compare with conventional algorithms using Anno robotic arm with Robot operating system (ROS), which provides solid and convenient testing environment for the operation of the arm.

After validating the algorithms on the "Randomly Piled Industrial Objects" dataset, the results of segmentation are listed in Table \ref{seg_result}. Improved DBSCAN is capable of segmenting point cloud of dense and randomly piled industrial objects, reducing noise and removing some bottom objects.
\par
\begin{figure}[htbp]
	\begin{minipage}[t]{0.45\linewidth}
	\centering
	\includegraphics[height=5cm,width=4cm]{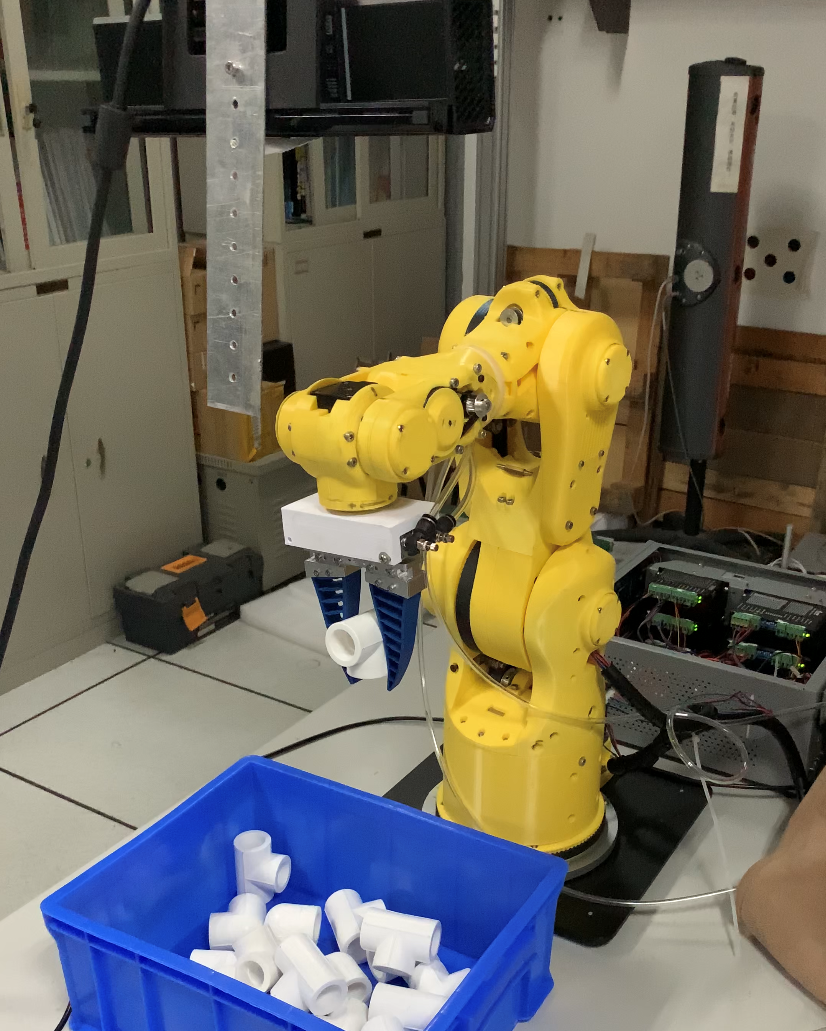}
	\caption{Grasp object on top.}
	\label{grasp}
	\end{minipage}
	\begin{minipage}[t]{0.45\linewidth}
	\centering
	\includegraphics[height=5cm,width=4cm]{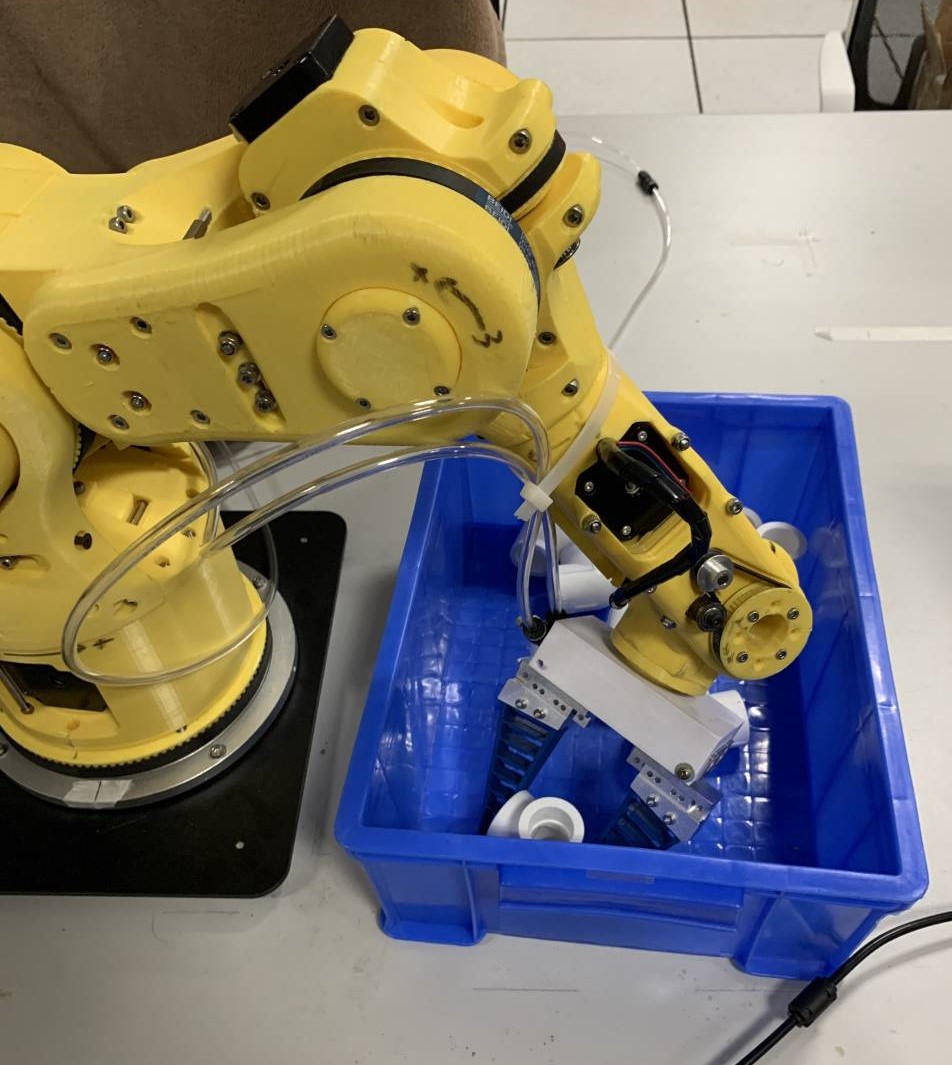}
	\caption{Grasp object in corner}
	\label{grasp2}
	\end{minipage}
	\end{figure}

Our grasp system is deployed to an Anno robotic arm with a kinect V2 camera, as shown in fig. \ref{grasp}. Two main factors, grasp-success-rate and collapse-rate, works to evaluate the performance of our method. Former is the ratio of the number of objects successfully moved out of the box to the total number of the objects, while the latter is the times of collisions to the total times of grasping. The results of the experiment with the proposed and comparative pipelines are listed in Table \ref{result}. Among the approaches involved in the experiments, our method gains the best performance.

\begin{table}[]
\begin{center}

	\caption{Segmentation results.}\label{seg_result}
	\begin{tabular}{@{}ccc@{}}
	\toprule
	Segmentation   & Accuracy & Runtime (s) \\ \midrule
	Euclidean Clustering         &  73\%       & 0.112          \\
	Region Growing        & 47\%         & 0.339          \\
	DBSCAN           & 80\%       & 0.354          \\	
	Our Method              & 95\%         & 0.084            \\ \bottomrule
	\end{tabular}
	\end{center}
\end{table}

\begin{table}[]
	\caption{Experiments results.}\label{result}
	\begin{tabular}{@{}cccc@{}}
	\toprule
	Segmentation   & Registration  & Success Rate & Collision Rate \\ \midrule
	Euclidean Clustering        & SCAIA+ICP     & 78\%         & 28\%           \\
	Euclidean Clustering        & NDT+ICP       & 71\%         & 31\%           \\
	Region Growing         & SCAIA+ICP     & 61\%         & 40\%           \\
	Region Growing         & NDT+ICP       & 56\%         & 35\%           \\
	\multicolumn{2}{c}{Our Method} & 92\%         & 7\%            \\ \bottomrule
	\end{tabular}
\end{table}

Photograph of the working robot is shown in Fig.\ref{grasp}. The gesture of the robot when it's grasping the object in corner is particularly presented in Fig.\ref{grasp2}.
\section{Conclusion}

In this paper, a pipeline for grasping random piled and dense objects fast, adaptively and safely is proposed. Our methods start at segmentate point cloud with modified DBSCAN-based algorithm, which is improved by combining the Region Growing algorithm and using Octree to speed up the calculation. Then, for coarse registration, PCA algorithm is used in our method and for fine registration, ICP is applied. We proposed Grasp Risk Score to evaluate each object by the collision probability, the stability of object and the whole system's stability. Through real test with Anno robot, our method is verified to be advanced in speed and robustness.

\bibliographystyle{IEEEtran}
\bibliography{refer}

\begin{thebibliography}{10}
\providecommand{\url}[1]{#1}
\csname url@samestyle\endcsname
\providecommand{\newblock}{\relax}
\providecommand{\bibinfo}[2]{#2}
\providecommand{\BIBentrySTDinterwordspacing}{\spaceskip=0pt\relax}
\providecommand{\BIBentryALTinterwordstretchfactor}{4}
\providecommand{\BIBentryALTinterwordspacing}{\spaceskip=\fontdimen2\font plus
\BIBentryALTinterwordstretchfactor\fontdimen3\font minus
  \fontdimen4\font\relax}
\providecommand{\BIBforeignlanguage}[2]{{%
\expandafter\ifx\csname l@#1\endcsname\relax
\typeout{** WARNING: IEEEtran.bst: No hyphenation pattern has been}%
\typeout{** loaded for the language `#1'. Using the pattern for}%
\typeout{** the default language instead.}%
\else
\language=\csname l@#1\endcsname
\fi
#2}}
\providecommand{\BIBdecl}{\relax}
\BIBdecl

\bibitem{oehler2011efficient}
B.~Oehler, J.~Stueckler, J.~Welle, D.~Schulz, and S.~Behnke, ``Efficient
  multi-resolution plane segmentation of 3d point clouds,'' in
  \emph{International Conference on Intelligent Robotics and
  Applications}.\hskip 1em plus 0.5em minus 0.4em\relax Springer, 2011, pp.
  145--156.

\bibitem{li2017improved}
L.~Li, F.~Yang, H.~Zhu, D.~Li, Y.~Li, and L.~Tang, ``An improved ransac for 3d
  point cloud plane segmentation based on normal distribution transformation
  cells,'' \emph{Remote Sensing}, vol.~9, no.~5, p. 433, 2017.

\bibitem{han2017review}
X.-F. Han, J.~S. Jin, M.-J. Wang, W.~Jiang, L.~Gao, and L.~Xiao, ``A review of
  algorithms for filtering the 3d point cloud,'' \emph{Signal Processing: Image
  Communication}, vol.~57, pp. 103--112, 2017.

\bibitem{narvaez2006point}
E.~A.~L. Narv{\'a}ez and N.~E.~L. Narv{\'a}ez, ``Point cloud denoising using
  robust principal component analysis.'' in \emph{GRAPP}, 2006, pp. 51--58.

\bibitem{schall2005robust}
O.~Schall, A.~Belyaev, and H.-P. Seidel, ``Robust filtering of noisy scattered
  point data,'' in \emph{Proceedings Eurographics/IEEE VGTC Symposium
  Point-Based Graphics, 2005.}\hskip 1em plus 0.5em minus 0.4em\relax IEEE,
  2005, pp. 71--144.

\bibitem{hu2006mean}
G.~Hu, Q.~Peng, and A.~R. Forrest, ``Mean shift denoising of point-sampled
  surfaces,'' \emph{The Visual Computer}, vol.~22, no.~3, pp. 147--157, 2006.

\bibitem{wang2013consolidation}
J.~Wang, K.~Xu, L.~Liu, J.~Cao, S.~Liu, Z.~Yu, and X.~D. Gu, ``Consolidation of
  low-quality point clouds from outdoor scenes,'' in \emph{Computer Graphics
  Forum}, vol.~32, no.~5.\hskip 1em plus 0.5em minus 0.4em\relax Wiley Online
  Library, 2013, pp. 207--216.

\bibitem{ye2011accurate}
M.~Ye, X.~Wang, R.~Yang, L.~Ren, and M.~Pollefeys, ``Accurate 3d pose
  estimation from a single depth image,'' in \emph{2011 International
  Conference on Computer Vision}.\hskip 1em plus 0.5em minus 0.4em\relax IEEE,
  2011, pp. 731--738.

\bibitem{taubin1995estimating}
G.~Taubin, ``Estimating the tensor of curvature of a surface from a polyhedral
  approximation,'' in \emph{Proceedings of IEEE International Conference on
  Computer Vision}.\hskip 1em plus 0.5em minus 0.4em\relax IEEE, 1995, pp.
  902--907.

\bibitem{xiao2006dynamic}
C.~Xiao, Y.~Miao, S.~Liu, and Q.~Peng, ``A dynamic balanced flow for filtering
  point-sampled geometry,'' \emph{The Visual Computer}, vol.~22, no.~3, pp.
  210--219, 2006.

\bibitem{chen2017point}
X.~Chen, Y.~Zhu, T.~Wu, and Z.~WANG, ``The point cloud registration technology
  based on sac-ia and improved icp,'' \emph{J Xi'an Polytech Univ}, vol.~31,
  no.~3, pp. 395--401, 2017.

\bibitem{rusu2009fast}
R.~B. Rusu, N.~Blodow, and M.~Beetz, ``Fast point feature histograms (fpfh) for
  3d registration,'' in \emph{2009 IEEE international conference on robotics
  and automation}.\hskip 1em plus 0.5em minus 0.4em\relax IEEE, 2009, pp.
  3212--3217.

\bibitem{zhang2019frequency}
S.~Zhang, H.~Wang, J.-g. Gao, and C.-q. Xing, ``Frequency domain point cloud
  registration based on the fourier transform,'' \emph{Journal of Visual
  Communication and Image Representation}, vol.~61, pp. 170--177, 2019.

\bibitem{li2015modified}
W.~Li and P.~Song, ``A modified icp algorithm based on dynamic adjustment
  factor for registration of point cloud and cad model,'' \emph{Pattern
  Recognition Letters}, vol.~65, pp. 88--94, 2015.

\bibitem{zapata2017using}
B.~S. Zapata-Impata, C.~Mateo~Agull{\'o}, P.~Gil, and J.~Pomares, ``Using
  geometry to detect grasping points on 3d unknown point cloud,'' 2017.

\bibitem{yan2020fast}
W.~Yan, Z.~Xu, X.~Zhou, Q.~Su, S.~Li, and H.~Wu, ``Fast object pose estimation
  using adaptive threshold for bin-picking,'' \emph{IEEE Access}, vol.~8, pp.
  63\,055--63\,064, 2020.

\bibitem{buchholz2010ransam}
D.~Buchholz, S.~Winkelbach, and F.~M. Wahl, ``Ransam for industrial
  bin-picking,'' in \emph{ISR 2010 (41st International Symposium on Robotics)
  and ROBOTIK 2010 (6th German Conference on Robotics)}.\hskip 1em plus 0.5em
  minus 0.4em\relax VDE, 2010, pp. 1--6.

\bibitem{buchholz2013efficient}
D.~Buchholz, M.~Futterlieb, S.~Winkelbach, and F.~M. Wahl, ``Efficient
  bin-picking and grasp planning based on depth data,'' in \emph{2013 IEEE
  International Conference on Robotics and Automation}.\hskip 1em plus 0.5em
  minus 0.4em\relax IEEE, 2013, pp. 3245--3250.

\bibitem{pochyly2017robotic}
A.~Pochyly, T.~Kubela, V.~Singule, and P.~{\v{C}}ih{\'a}k, ``Robotic
  bin-picking system based on a revolving vision system,'' in \emph{2017 19th
  International Conference on Electrical Drives and Power Electronics
  (EDPE)}.\hskip 1em plus 0.5em minus 0.4em\relax IEEE, 2017, pp. 347--352.

\bibitem{martinez2015automated2s}
C.~Martinez, H.~Chen, and R.~Boca, ``Automated 3d vision guided bin picking
  process for randomly located industrial parts,'' in \emph{2015 IEEE
  International Conference on Industrial Technology (ICIT)}.\hskip 1em plus
  0.5em minus 0.4em\relax IEEE, 2015, pp. 3172--3177.

\bibitem{chang2016eye}
W.-C. Chang and C.-H. Wu, ``Eye-in-hand vision-based robotic bin-picking with
  active laser projection,'' \emph{The International Journal of Advanced
  Manufacturing Technology}, vol.~85, no. 9-12, pp. 2873--2885, 2016.

\bibitem{kim2020bin}
K.-K. Kim, S.-S. Kang, J.-Y. Lee, J.-H. Kim, J.-B. Kim, and S.~Sung-Woong,
  ``Bin-picking system and method for bin-picking,'' Jan.~21 2020, uS Patent
  10,540,567.

\bibitem{he2020inward}
Z.~He, H.~Wang, X.~Zhao, S.~Zhang, and J.~Tan, ``Inward-region-growing-based
  accurate partitioning of closely stacked objects for bin-picking,''
  \emph{Measurement Science and Technology}, 2020.

\bibitem{schnabel2007efficient}
R.~Schnabel, R.~Wahl, and R.~Klein, ``Efficient ransac for point-cloud shape
  detection,'' in \emph{Computer graphics forum}, vol.~26, no.~2.\hskip 1em
  plus 0.5em minus 0.4em\relax Wiley Online Library, 2007, pp. 214--226.

\bibitem{ester1996density}
M.~Ester, H.-P. Kriegel, J.~Sander, X.~Xu \emph{et~al.}, ``A density-based
  algorithm for discovering clusters in large spatial databases with noise.''
  in \emph{Kdd}, vol.~96, no.~34, 1996, pp. 226--231.

\bibitem{han2018towards}
S.~Han, ``Towards efficient implementation of an octree for a large 3d point
  cloud,'' \emph{Sensors}, vol.~18, no.~12, p. 4398, 2018.

\bibitem{rabbani2006segmentation}
T.~Rabbani, F.~Van Den~Heuvel, and G.~Vosselmann, ``Segmentation of point
  clouds using smoothness constraint,'' \emph{International archives of
  photogrammetry, remote sensing and spatial information sciences}, vol.~36,
  no.~5, pp. 248--253, 2006.

\bibitem{yuan20163d}
C.~Yuan, X.~Yu, and Z.~Luo, ``3d point cloud matching based on principal
  component analysis and iterative closest point algorithm,'' in \emph{2016
  International Conference on Audio, Language and Image Processing
  (ICALIP)}.\hskip 1em plus 0.5em minus 0.4em\relax IEEE, 2016, pp. 404--408.

\bibitem{kozai2018determining}
K.~Kozai and M.~Hashimoto, ``Determining robot grasping-parameters by
  estimating" picking risk",'' in \emph{2018 International Workshop on Advanced
  Image Technology (IWAIT)}.\hskip 1em plus 0.5em minus 0.4em\relax IEEE, 2018,
  pp. 1--4.

\end{thebibliography}

\vspace{12pt}

\end{document}